# Adaptive Online Learning with Regularized Kernel for One-class Classification

Chandan Gautam*, Aruna Tiwari, Sundaram Suresh, *Senior Member, IEEE,* and Kapil Ahuja

*Abstract*—This paper presents adaptive Online learning with Regularized Kernel-based One-Class Extreme Learning Machine (ELM) classifiers for detection of outliers, and are collectively referred to as "*ORK-OCELM*". Two frameworks viz., boundary and reconstruction, are presented to detect the target class in *ORK-OCELM*. The kernel hyperplane based baseline one-class ELM model considers whole data in a single chunk, however, the proposed one-class classifiers are adapted in an online fashion from the stream of training samples.

The performance of *ORK-OCELM* is evaluated on standard benchmark as well as synthetic datasets for both types of environment, i.e. stationary and non-stationary. While evaluating on stationary datasets, these classifiers are compared against batch learning based one-class classifiers. Similarly, while evaluating on non-stationary datasets, comparison is done with incremental learning based online one-class classifiers. The results indicate that the proposed classifiers yield better or similar outcomes for both. In the non-stationary dataset evaluation, adaptability of proposed classifiers in a changing environment is also demonstrated. It is further shown that proposed classifiers have large stream data handling capability even under limited system memory.

Moreover, the proposed classifiers gain significant time improvement compared to traditional online one-class classifiers (in all aspects of training and testing). A faster learning ability of the proposed classifiers makes them more suitable for real-time anomaly detection.

## I. INTRODUCTION

THE term 'One-class classification'(OCC) was coined by Moya et al. [1]. The OCC has been generally employed to solve the problem of novelty, outlier, intrusion, anomaly or fault detection [2]. These detection techniques are applicable for various types of applications across many disciplines [3]–[10]. Such types of problems can also be solved by multi-class classification when samples of both the classes, the normal and the outlier class, are available [11]–[15]. However, the OCC can be applied even when either the data of only one-class is available or the data belonging to other class is very rare or very expensive to collect. A one-class classifier simply describes the data, and therefore, is sometimes also referred to as a data descriptor. One-class classifiers are generally divided into two parts [2] viz., kernel and non-kernel-based one-class classifiers. Mostly, kernel-based methods have performed better than non-kernel methods for OCC tasks [2].

In this and the next paragraph, we briefly revisit OCC for both offline and online learning. Most researchers have developed kernel-based OCC for offline learning and have considered either Support Vector Machine (SVM) or Kernel Principal Component Analysis (KPCA) as a base classifier. Two types of SVM based one-class classifiers have been developed for offline learning viz., One-Class SVM (OCSVM) [16] and Support Vector Domain Description (SVDD) [17].

Next, we briefly describe the two approaches to update the model when going from offline to online i.e. batch re-computation and incremental update. In batch re-computation, the old model is completely discarded and a new model is constructed based upon the current set of training samples, which may include certain obsolete or old samples (single sample or block of samples). However, an incremental update is performed in two steps. First, the current model is updated by forgetting certain obsolete or old samples. Then, the current model is retrained by only the new upcoming samples with the help of the updated model (generated by the forgetting mechanism).

Online learning has attracted researchers in recent years due to its capability to handle high volume of streaming data with less computational and storage costs [18]–[21]. Similar as offline learning, online learning based one-class classifiers have also been mainly derived from two of the kernel-based classifiers, i.e. SVM and KPCA. SVDD based incremental one-class classifier (incSVDD) for online learning has been developed by Tax and Laskov [22] and further explored by other researchers [18], [19]. KPCA based online one-class classifier (OKPCA) has been developed by Chatzigiannakis and Papavassiliou [23], where the model is constructed in a batch mode for all new samples. Later, an incremental way of model construction for KPCA has been developed for new samples [20], [21].

Above discussed kernel-based one-class classifiers use an iterative approach, which is very time consuming. This issue has been addressed by Single Layer Feed-forward Network ($SLFN$) in the past few decades. $SLFN$ got a substantial attention from researchers for multi-class classification and regression task due to its fast training capability. Huang et al. [28] [29] have also developed a $SLFN$, called as Extreme Learning Machine (ELM). ELM loosely connects machine learning techniques with biological learning mechanisms [30] [31] and hence, has attracted researcher towards it.

*Corresponding Author

Chandan Gautam is with the Indian Institute of Technology Indore, (e-mail: chandangautam31@gmail.com).

Aruna Tiwari is with the Indian Institute of Technology Indore, (e-mail: artiwari@iiti.ac.in).

Sundaram Suresh is with the Nanyang Technological University, Singapore, (e-mail: ssundaram@ntu.edu.sg).

Kapil Ahuja is with the Indian Institute of Technology Indore, (e-mail: kahuja@iiti.ac.in).

TABLE I: Different variants of existing and proposed ELM-based one-class classifier

| | Learning Type | | Feature Mapping Type | | Forgetting Mechanism | Environment Type | |
|---|---|---|---|---|---|---|---|
| | Offline | Online | Random | Kernel | | Stationary | Non-stationary |
| [24] | ✓ | ✗ | ✓ | ✓ | ✗ | ✓ | ✗ |
| [25] | ✓ | ✗ | ✓ | ✓ | ✗ | ✓ | ✗ |
| [26], [27] | ✓ | ✓ | ✓ | ✗ | ✗ | ✓ | ✗ |
| **Proposed** | ✓[1] | ✓ | ✗ | ✓ | ✓ | ✓ | ✓ |

[1] Here, the offline classifier can be treated as a special case of the online classifier.

In the past few decades, ELM has been well expanded in the two common dimensions; theory [32]–[35] and application [36]–[41]. Various variants of ELM have been proposed, e.g., bayesian ELM [42], sparse bayesian ELM [43], semi-supervised and unsupervised ELM [44], incremental ELM [45], [46], weighted ELM [47], multi-layer ELM [48]–[50], domain adaption transfer learning ELM [51] etc. All the above discussed variants are for multi-class classification and offline learning. Online version of ELM has also been developed for multi-class classification and regression tasks with random [52], [53] and kernel feature mapping [54], [55]. A detailed survey on ELM and its applications can be found in [56].

Recently, Leng et al. [24] have developed a basic ELM for OCC task (One-class Kernelized ELM, i.e. OCKELM), which supports only offline learning. Later, OCKELM is improvised by Iosifidis et al. [25] and Gautam et al. [27]. Iosifidis et al. [25] have proposed ELM based one-class classifier that uses geometric information of a class for offline learning. Gautam et al. [26] [27] have also explored ELM based one-class classifier mainly for offline learning using Auto-associative Kernelized ELM (AAKELM). The results in [27] indicate that the kernel feature mapping based offline methods outperform the random feature mapping based offline methods. In [27], authors have also developed random feature mapping based online one-class classifiers, which exhibited very inferior performance compared to their corresponding kernel-based offline one-class classifiers. Therefore, this paper presents ELM based online one-class classifiers with kernel feature mapping, which performs similar to their corresponding kernel-based offline one-class classifiers. Comparison of different variants of ELM based one-class classifiers and this proposed work is given in Table I.

The main **contributions** (see Table I) of this work are listed below:

(i) This paper introduces kernel feature mapping based ELM for OCC in online mode, which constructs the model with the incremental approach and is competitive to OCC in offline mode.
(ii) Proposed ELM based one-class classifiers are adaptive in nature, and therefore, are capable of handling large streams of data in both types of environments, i.e. stationary and non-stationary.
(iii) As data with incremental drift are more challenging to classify, and hence, the proposed classifiers are developed to handle such types of drift. Moreover, these are tested on various types of non-stationary environments. As a use case, data from normal and the outlier classes are distributed through unimodal or multimodal Gaussian distributions etc.
(iv) Proposed classifiers are developed for two types of frameworks; boundary and reconstruction. The boundary framework based method is developed as a fast online single output node architecture. The reconstruction based approach is developed as a fast online autoencoder.

The rest of the paper is organized as follows: Section II discusses the proposed work. Performance evaluation is discussed in Section III. Section IV contains conclusion and future work.

## II. PROPOSED WORK

In this section, ELM based online sequential one-class classifier is modeled by taking both factors viz., regularization and kernelization, into account. This is called as **O**nline learning with **R**egularized **K**ernel for **O**ne-**C**lass ELM (*ORK-OCELM*). Regularization factor helps classifier to achieve a better generalization capability for noisy data. Tikhonov regularization [57] has been employed due to its capability to handle ill-posed and singular problems that generally emerge in solving inverse problems. Following subsections discusses *ORK-OCELM* for two types of framework viz., **B**oundary - *ORK-OCELM(B)* and **R**econstruction - *ORK-OCELM(R)*:

### A. ORK-OCELM(B): Boundary Framework Based Approach

In Boundary framework based *ORK-OCELM* i.e. *ORK-OCELM(B)*, model is trained by only target data $X$ and approximates all data to a real number. Fig. 1 shows online OCC with single output node architecture. Here, given a stream of training data $X$: $\{(x_1, c_1), (x_2, c_2), ..., (x_t, c_t), ...\}$, where $x_t = [x_t^1, x_t^2, ..., x_t^n] \in \Re^n$ is $n$-dimensional input of the $t^{th}$ sample, and $c_t$ is the class label of the target class, which is same for all of training data. Input layer that takes data for $t^{th}$ input sample is coded as $(x_t, R_t)$ because model has to approximate all data to a real number $R$. Target output vector $R$ is represented as $[R_1, R_2....R_t, ...]$, however, value of $R_t$ will be same for all samples. Here, value of $R_t$ is considered as 1 for all experiments. Further, kernel feature mapping has been employed between input and hidden layer. Here, kernel matrix is represented as $\Phi = H^T H$, where $H$ denotes random feature mapping. Note that $\Phi$ and $H$ can also be written in terms of $X$, i.e. $\Phi = \phi(X) = K(X, X)$ and $H = h(X)$, where $\phi$ and $h$ are a function for random and kernel feature mapping, and $K$ denotes inner product. Here, hidden layer output i.e. kernel matrix $\Phi$ is a square symmetric matrix of size $t \times t$. Output weight vector $\boldsymbol{\beta}$ for any $t$ samples in sequence of data is represented as $[\beta_{11}, \beta_{21}....\beta_{(t-1)1}, \beta_{tt}]$. $\widehat{\boldsymbol{R}} = [\widehat{R}_1, \widehat{R}_2...\widehat{R}_t, ...]$ is the predicted output vector and $\widehat{R}_t$ is the predicted output for $t^{th}$ sample. $\widehat{\boldsymbol{c}} = [\widehat{c}_1, \widehat{c}_2...\widehat{c}_t, ...]$ is the predicted class vector, where, $\widehat{c}_t$ is the predicted class for $t^{th}$ sample. Based on these preliminaries, boundary based online learning algorithm **ORK-OCELM(B)** is discussed below in **four phases** i.e. **Initialization, Update, Forgetting** and **Decision**.

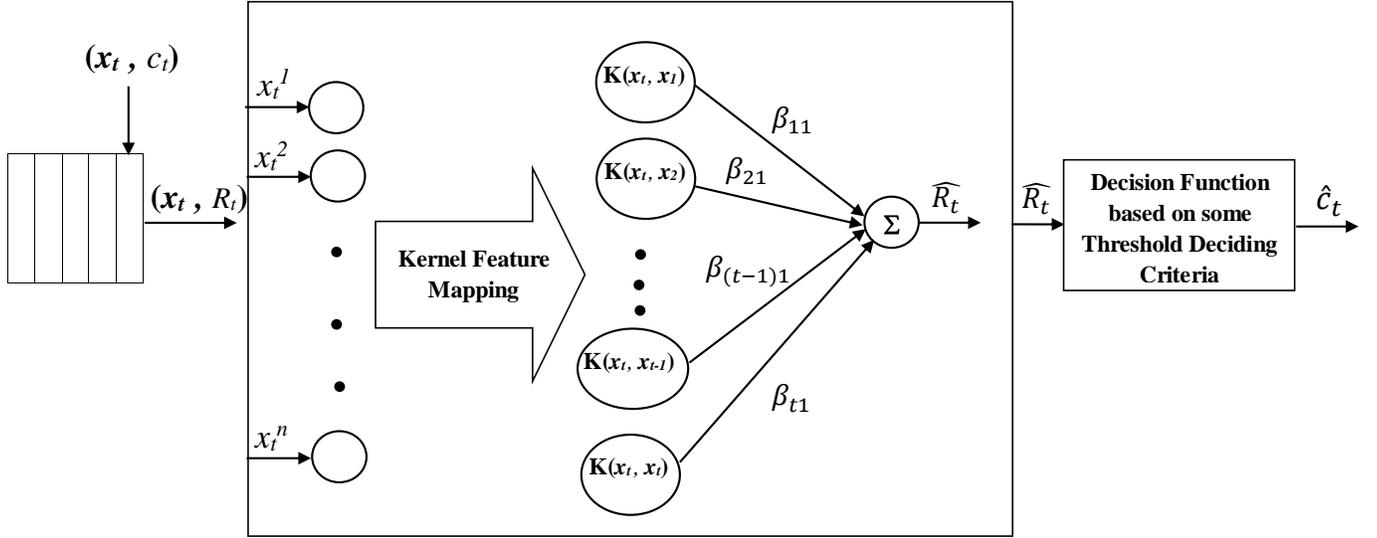

Fig. 1: Schematic Diagram of Online Sequential One-class Extreme Learning Machine: Boundary Based

*1) Initialization Phase:* In this phase, first, basic ELM formulation is derived for random feature mapping with hidden layer matrix $\boldsymbol{H}$ and later, this formulation is derived for kernel feature mapping with kernel matrix $\boldsymbol{\Phi}$ using Representer Theorem [58].

For the initial chunk $\{\boldsymbol{X_0}, \boldsymbol{R_0}\}$ of size $N_0$, the objective is to minimize both output weight vector $\boldsymbol{w_0}$, it is output weight vector for random feature mapping case, as well as error $\boldsymbol{E_0}$ between expected $(\boldsymbol{R_0})$ and predicted output $(\boldsymbol{w_0^T H_0})$. Minimization problem for random feature mapping case can be written as follows:

$$\text{Minimize}: L = \frac{1}{2}\|\boldsymbol{w_0}\|^2 + \lambda \frac{1}{2}\|\boldsymbol{E_0}\|^2 \qquad (1)$$
$$\text{Subject to}: \boldsymbol{w_0^T H_0} = \boldsymbol{R_0} - \boldsymbol{E_0}$$

Here, $\lambda$ is used as a regularization parameter.

Representer Theorem [58] is exploited in (2), which describes the weight vector $\boldsymbol{w_o}$ as a linear combination of the training data representation in ELM space $(\boldsymbol{H_0})$ and a reconstruction weight vector $(\boldsymbol{\beta_0})$, it is output weight vector for the kernel feature mapping case, as follows:

$$\boldsymbol{w_0} = \boldsymbol{H_0}\boldsymbol{\beta_0}. \qquad (2)$$

Further, by using Representer theorem, minimization problem in (1) can be reformulated as follows:

$$\text{Minimize}: L = \frac{1}{2}\boldsymbol{\beta_0^T H_0^T H_0 \beta_0} + \lambda \frac{1}{2}\|\boldsymbol{E_0}\|^2 \qquad (3)$$
$$\text{Subject to}: \boldsymbol{\beta_0^T H_0^T H_0} = \boldsymbol{R_0} - \boldsymbol{E_0}$$

Now, substituting kernel matrix of initial training samples, $\boldsymbol{\Phi_0} = \phi(\boldsymbol{X_0}) = (\boldsymbol{H_0})^T \boldsymbol{H_0} = K(\boldsymbol{X_0}, \boldsymbol{X_0})$, in the above equation and obtain the following minimization problem:

$$\text{Minimize}: L = \frac{1}{2}\boldsymbol{\beta_0^T \Phi_0 \beta_0} + \lambda \frac{1}{2}\|\boldsymbol{E_0}\|^2 \qquad (4)$$
$$\text{Subject to}: \boldsymbol{\beta_0^T \Phi_0} = \boldsymbol{R_0} - \boldsymbol{E_0}$$

By using Lagrangian relaxation [59], (4) can be written as dual optimization problem:

$$\text{Minimize}: L = \frac{1}{2}\boldsymbol{\beta_0^T \Phi_0 \beta_0} + \lambda \frac{1}{2}\|\boldsymbol{E_0}\|^2 \qquad (5)$$
$$- \alpha(\boldsymbol{\beta_0^T \Phi_0} - \boldsymbol{R_0} + \boldsymbol{E_0})$$

where $\alpha$ is the Lagrange multiplier, which is employed to combine the constraint with the minimization problem. By using Karush-Kuhn-Tucker (KKT) theorem [60], take partial derivatives of (5) with respect to all variables $\boldsymbol{\beta_0}, \boldsymbol{E_0}$ and $\alpha$. That is

$$\frac{\partial L}{\partial \boldsymbol{\beta_0}} = 0 \Rightarrow \boldsymbol{\beta_0} = \alpha$$
$$\frac{\partial L}{\partial \boldsymbol{E_0}} = 0 \Rightarrow \lambda \boldsymbol{E_0} = \alpha \qquad (6)$$
$$\frac{\partial L}{\partial \alpha} = 0 \Rightarrow \boldsymbol{\beta_0^T \Phi_0} - \boldsymbol{R_0} + \boldsymbol{E_0} = 0$$

After solving the above set of equations in (6) for $\boldsymbol{\beta_0}$, it is obtained as follows:

$$\boldsymbol{\beta_0} = (\boldsymbol{\Phi_0} + \frac{1}{\lambda}\boldsymbol{I})^{-1}\boldsymbol{R_0} \qquad (7)$$

Here, $\boldsymbol{I}$ is an identity matrix.

Finally, $\boldsymbol{\beta_0}$ for initial training samples is obtained as follows:

$$\boldsymbol{\beta_0} = \boldsymbol{P_0 R_0}, \text{whrere } \boldsymbol{P_0} = \left(\boldsymbol{\Phi_0} + \frac{1}{\lambda}\boldsymbol{I}\right)^{-1} \qquad (8)$$

$\boldsymbol{\Phi_0}$ is defined based on Mercer's condition. That is, any kernel method which satisfies Mercer's condition can be adopted as the kernel for the classifier. Obtained $\boldsymbol{\Phi_0}$ and $\boldsymbol{P_0}$ are stored in different variables $\boldsymbol{\Phi}$ and $\boldsymbol{P}$ respectively as these variables need to be updated when new training samples arrives. Update of $\boldsymbol{\Phi}$ and $\boldsymbol{P}$ are discussed in the second phase. $\boldsymbol{\Phi}$, $\boldsymbol{P}$, $\boldsymbol{\Phi_0}$ and $\boldsymbol{P_0}$ are defined as follows:

$$\boldsymbol{\Phi} = \left(\boldsymbol{\Phi_0} + \frac{1}{\lambda}\boldsymbol{I}\right) = \left(\begin{bmatrix} \Omega_{11} & \Omega_{12} & \ldots & \Omega_{1N_0} \\ \Omega_{21} & \Omega_{22} & \ldots & \Omega_{2N_0} \\ \vdots & \vdots & \ddots & \vdots \\ \Omega_{N_0 1} & \Omega_{N_0 2} & \ldots & \Omega_{N_0 N_0} \end{bmatrix} + \frac{1}{\lambda}\boldsymbol{I}\right) \quad (9)$$

$$\boldsymbol{P} = \boldsymbol{P_0} = \boldsymbol{\Phi}^{-1} = \left(\begin{bmatrix} \Omega_{11} & \Omega_{12} & \ldots & \Omega_{1N_0} \\ \Omega_{21} & \Omega_{22} & \ldots & \Omega_{2N_0} \\ \vdots & \vdots & \ddots & \vdots \\ \Omega_{N_0 1} & \Omega_{N_0 2} & \ldots & \Omega_{N_0 N_0} \end{bmatrix} + \frac{1}{\lambda}\boldsymbol{I}\right)^{-1} \quad (10)$$

*2) Update $\boldsymbol{P}$ and $\boldsymbol{\Phi}$:* Initially, $\boldsymbol{\Phi}$ and $\boldsymbol{P}$ are calculated as per (9) and (10). Here, $\boldsymbol{\Phi}$ represents kernel matrix and $\boldsymbol{P}$ represents inverse of this kernel matrix for all the arrived samples until now for training. $\boldsymbol{\Phi}$ and $\boldsymbol{P}$ are updated continuously for any upcoming new samples $\boldsymbol{X^v}$ as per (11), where, $\boldsymbol{X^v} = \{(\boldsymbol{x}_1^v, c_1), (\boldsymbol{x}_2^v, c_2), ..., (\boldsymbol{x}_s^v, c_s)\}$ and $\boldsymbol{X^v} \subset \boldsymbol{X}$. Current value of $\boldsymbol{\Phi}$ is represented as $\boldsymbol{\Phi_u}$, which is generated by using old samples $\boldsymbol{X^u} \subset \boldsymbol{X}$. Now, $\boldsymbol{\Phi}$ is calculated after arrival of the new sample as follows:

$$\boldsymbol{\Phi} = \begin{bmatrix} \boldsymbol{\Phi_u} & \boldsymbol{\Phi_{u,v}} \\ (\boldsymbol{\Phi_{u,v}})^T & \boldsymbol{\Phi_v} \end{bmatrix} \quad (11)$$

Here, $\boldsymbol{\Phi}$ is a combination of four block matrices. Block matrix $\boldsymbol{\Phi_{u,v}}$ and $\boldsymbol{\Phi_v}$ above are calculated as per (12), which is discussed next. Let the number of samples processed until now be $b$ and number of samples in the current chunk be $s$. $b$ is initially equal to $N_0$. Update $b$ and $s$ each time when calculation starts for new samples. The block matrices $\boldsymbol{\Phi_v}$ and $\boldsymbol{\Phi_{u,v}}$ are defined as follows:

$$\boldsymbol{\Phi_v} = \left(\begin{bmatrix} K(x_1^v, x_1^v) & \ldots & K(x_1^v, x_s^v) \\ \vdots & \ddots & \vdots \\ K(x_s^v, x_1^v) & \ldots & K(x_s^v, x_s^v) \end{bmatrix} + \frac{1}{\lambda}\boldsymbol{I}\right) \quad (12)$$

$$\boldsymbol{\Phi_{u,v}} = \begin{bmatrix} K(x_1^u, x_1^v) & \ldots & K(x_1^u, x_s^v) \\ \vdots & \ddots & \vdots \\ K(x_b^u, x_1^v) & \ldots & K(x_b^u, x_s^v) \end{bmatrix} \quad (13)$$

$\boldsymbol{P}$, which is inverse of $\boldsymbol{\Phi}$ given in (11), is defined as follows:

$$\boldsymbol{P} = \boldsymbol{\Phi}^{-1} = \begin{bmatrix} \boldsymbol{\Phi_u} & \boldsymbol{\Phi_{u,v}} \\ (\boldsymbol{\Phi_{u,v}})^T & \boldsymbol{\Phi_v} \end{bmatrix}^{-1} \quad (14)$$

Further, compute the inverse in (14) using block matrix inverse [61] given by:

$$\boldsymbol{S} = \boldsymbol{D}^{-1} = \begin{bmatrix} D_{11} & D_{12} \\ D_{21} & D_{22} \end{bmatrix}^{-1} = \begin{bmatrix} S_{11} & S_{12} \\ S_{21} & S_{22} \end{bmatrix} \quad (15)$$

$\boldsymbol{S}$ in (15) can be written as follows to obtain inverse:

$$\begin{bmatrix} D_{11} & D_{12} \\ D_{21} & D_{22} \end{bmatrix} \begin{bmatrix} S_{11} & S_{12} \\ S_{21} & S_{22} \end{bmatrix} = \begin{bmatrix} I & 0 \\ 0 & I \end{bmatrix} \quad (16)$$

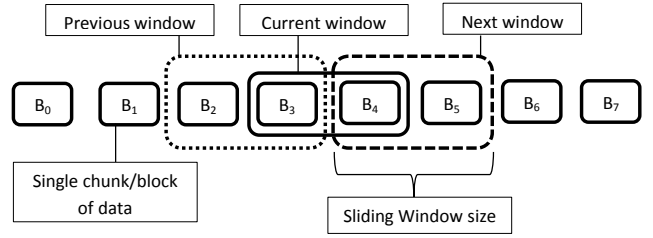

Fig. 2: Illustration of Sliding Window for a Given Data Stream

where, $S_{21} = S_{12}^T$ and $D_{21} = D_{12}^T$.
After solving (16), the following solution is obtained:

$$S_{11} = (D_{11} - D_{12}D_{22}^{-1}D_{21})^{-1}$$
$$S_{12} = -D_{11}^{-1}D_{12}(D_{22} - D_{21}D_{11}^{-1}D_{12})^{-1}$$
$$S_{21} = -D_{22}^{-1}D_{21}(D_{11} - D_{12}D_{22}^{-1}D_{21})^{-1}$$
$$S_{22} = (D_{22} - D_{21}D_{11}^{-1}D_{12})^{-1}$$

Hence, (14) can be rewritten as follows:

$$\boldsymbol{P} = \boldsymbol{\Phi}^{-1} = \begin{bmatrix} \boldsymbol{\Phi_u} & \boldsymbol{\Phi_{u,v}} \\ (\boldsymbol{\Phi_{u,v}})^T & \boldsymbol{\Phi_v} \end{bmatrix}^{-1} = \begin{bmatrix} P_{11} & P_{12} \\ P_{21} & P_{22} \end{bmatrix} \quad (17)$$

Based on the above discussion, $P_{11}$, $P_{12}$, $P_{21}$ and $P_{22}$ in (17) are given as below:

$$P_{11} = (\boldsymbol{\Phi_u} - \boldsymbol{\Phi_{u,v}} \boldsymbol{\Phi_v}^{-1} \boldsymbol{\Phi_{u,v}}^T)^{-1}$$
$$P_{12} = -\boldsymbol{\Phi_u}^{-1} \boldsymbol{\Phi_{u,v}} P_{22}$$
$$P_{21} = -\boldsymbol{\Phi_v}^{-1} \boldsymbol{\Phi_{u,v}}^T P_{11} \quad (18)$$
$$P_{22} = (\boldsymbol{\Phi_v} - \boldsymbol{\Phi_{u,v}}^T \boldsymbol{\Phi_u}^{-1} \boldsymbol{\Phi_{u,v}})^{-1}$$

Further, $P_{11}$ can be expanded by employing the Woodbury formula [62] as follows:

$$P_{11} = (\boldsymbol{\Phi_u} - \boldsymbol{\Phi_{u,v}} \boldsymbol{\Phi_v}^{-1} \boldsymbol{\Phi_{u,v}}^T)^{-1}$$
$$= \boldsymbol{\Phi_u}^{-1} - \boldsymbol{\Phi_u}^{-1} \boldsymbol{\Phi_{u,v}} (\boldsymbol{\Phi_{u,v}}^T \boldsymbol{\Phi_u}^{-1} \boldsymbol{\Phi_{u,v}} \quad (19)$$
$$+ \boldsymbol{\Phi_v}^{-1})^{-1} \boldsymbol{\Phi_{u,v}}^T \boldsymbol{\Phi_u}^{-1}$$

Woodbury formula [62] is employed instead of computing the direct inverse because now there is a need to compute two inverses, i.e. $\boldsymbol{\Phi_u}^{-1}$ and $\boldsymbol{\Phi_v}^{-1}$ independently, where $\boldsymbol{\Phi_u}^{-1}$ is already computed in the previous iteration. The advantage of this approach in terms of the time and storage complexity is described in Section III-E. $P_{12}$, $P_{21}$ and $P_{22}$ can be updated in a similar manner.

*3) Adaptive Learning:* During online learning, data increases continuously, and therefore, it creates two issues: how will the algorithm learn continuously if the distribution of training samples changes, and/or if the memory of system exhausts (as memory is not infinite). Both of these issues can be addressed by a forgetting mechanism with a sliding window as shown in Fig. 2. This mechanism unlearns the old or irrelevant samples by unlearning the trained model. Further, relearning on new samples can be done by online learning as discussed in the previous sections.

**Forgetting Mechanism for the proposed classifier:**
Suppose, $\boldsymbol{P_{curr}} = \boldsymbol{\Phi_{curr}^{-1}} \in \Re^{s \times s}$ is the inverse of the

current kernel matrix $\Phi_{curr} \in \Re^{s \times s}$. Now, the impact of learning of old $f$ samples needs to be removed. In this mechanism, two things viz., kernel matrix and the inverse of that kernel matrix, need to be updated before moving to learning of new samples. Modified kernel matrix $\phi_{new}$ can be simply generated by removing rows and columns of the corresponding samples from the current kernel matrix $\Phi_{curr}$. Modified inverse matrix $P_{new}$ is calculated after removing (forgetting) few samples from $\Phi_{curr}$ as follows:

$$P_{curr} = \Phi_{curr}^{-1} = \begin{bmatrix} F_{11} & F_{12} \\ F_{12}^T & R_{22} \end{bmatrix}^{-1} = \begin{bmatrix} Fi_{11} & Fi_{12} \\ Fi_{12}^T & Ri_{22} \end{bmatrix} \quad (20)$$

or

$$\begin{bmatrix} Fi_{11} & Fi_{12} \\ Fi_{12}^T & Ri_{22} \end{bmatrix} \begin{bmatrix} F_{11} & F_{12} \\ F_{12}^T & R_{22} \end{bmatrix} = \begin{bmatrix} I & 0 \\ 0 & I \end{bmatrix} \quad (21)$$

Suppose, if it is required to delete $F_{11}, F_{12}$ and $F_{12}^T$ from $\Phi_{curr}$ and to calculate the inverse of the remaining block $R_{22}$ by reusing the currently available inverse $P_{curr}$. Multiplying both sides of (21) with

$$\begin{bmatrix} I & 0 \\ -Fi_{12}^T Fi_{11}^{-1} & I \end{bmatrix} \quad (22)$$

and we get

$$\begin{bmatrix} Fi_{11} & Fi_{12} \\ 0 & Ri_{22} - Fi_{12}^T Fi_{11}^{-1} Fi_{12} \end{bmatrix} \begin{bmatrix} F_{11} & F_{12} \\ F_{12}^T & R_{22} \end{bmatrix} \\ = \begin{bmatrix} I & 0 \\ -Fi_{11}^{-1} Fi_{12}^T & I \end{bmatrix} \quad (23)$$

From (23), $P_{new}$ can be obtained after deletion of $F_{11}, F_{12}$ and $F_{12}^T$ from $\Phi_{curr}$ as follows:

$$\begin{aligned} \left( Ri_{22} - Fi_{12}^T Fi_{11}^{-1} Fi_{12} \right) R_{22} &= I \\ P_{new} = R_{22}^{-1} &= Ri_{22} - Fi_{12}^T Fi_{11}^{-1} Fi_{12} \end{aligned} \quad (24)$$

*4) Decision Phase:* After processing the training dataset $X_{curr}^{tr}$ in the current window, output function can be written for any set of $k$ samples $X_k = \{x_1, x_2, ..., x_k\}$ as follows:

$$f(X_k) = \begin{bmatrix} K(X_{curr}^{tr}, x_1) \\ \vdots \\ K(X_{curr}^{tr}, x_k) \end{bmatrix}^T PR \quad (25)$$

where, $K(X_{curr}^{tr}, x_i)$ denotes kernel vector for the $i^{th}$ sample $x_i$. Further, perform the following steps to decide whether any sample is outlier or not:

(i) Calculate distance ($d$) between the predicted value of the $t^{th}$ training sample and R as follows:

$$d(x_t) = |f(x_t) - R_t| = \left| \hat{R}_t - R_t \right| \quad (26)$$

(ii) After calculating distances $d$ as per (26), sort the differences in decreasing order. Further, reject few percent of training samples based on the deviation. Most deviated samples are rejected first because they are most probably far from distribution of the target data. The threshold ($\theta$) is decided based on these deviations as follows:

$$\theta = d(\lfloor \eta * N \rfloor) \quad (27)$$

---

**Algorithm 1** *ORK-OCELM(B)*:Boundary Based Approach

**Input:** Training set $X = (x_1, c_1), (x_2, c_2), ..., (x_{N_0}, c_{N_0}), ..., (x_t, c_t), ...$

**Output:** Whether each sample belongs to Target class or Outlier class

**Initialization Phase**

1: Pass initial set of samples $X_0, R_0$ to the classifier as: $\{(x_1, R_1), (x_2, R_2), ..., (x_{N_0}, R_{N_0})\}$
   // For first chunk of $N_0$ samples, follow the steps as below.
2: Employ kernel feature mapping: $\Phi_0 = \phi(X_0)$.
3: Output Weight $\beta_0$ for $(X_0, R_0)$:
$$\beta_0 \leftarrow P_0 R_0$$
$$P_0 \leftarrow \Phi_0^{-1}$$

**End of Initialization Phase**
// For second chunk onwards, following steps are required
$$\Phi \leftarrow \Phi_0$$
$$P \leftarrow P_0$$

4: **for** $i = 1$ to last chunk of data in $X$ **do**
5:   Set size of chunk at the current stage as $s$
6:   Update the final kernel matrix $\Phi$ and its inverse $P$ in two steps as given below:
     **Step 1: Forgetting Phase**
7:   Remove the impact of $s$ old samples from the current inverse $P$ using (24), i.e.,
     $$P_{new} = R_{22}^{-1} = Ri_{22} - Fi_{12}^T Fi_{11}^{-1} Fi_{12}$$
8:   Update the kernel matrix $\Phi$ by removing those rows and columns which were generated due to the old $s$ samples.
     **Step 2: Retraining Phase**
9:   Update the kernel matrix $\Phi$ as per (11), i.e.,
     $$\Phi = \begin{bmatrix} \Phi_u & \Phi_{u,v} \\ (\Phi_{u,v})^T & \Phi_v \end{bmatrix}$$
10:  Calculate $\Phi_v$ for $i^{th}$ chunk by using (12)
11:  Calculate $\Phi_{u,v}$ for $i^{th}$ chunk by using (13)
12:  Compute the inverse of updated kernel matrix $\Phi$, i.e., $P = \Phi^{-1}$ using block inverse as discussed in (15)-(19)
13:  $b = b + s$
14:  Update the Output Weight using the value of $R$ and updated $P$ as:
     $$\beta = PR$$
15:  Compute the predicted value by using output function $f(X_k)$, which is defined in (25)
16:  Calculate distances ($d$) between predicted value of training sample and $R$ as per (26), i.e.,
     $$d(x_t) = |f(x_t) - R_t| = \left| \hat{R}_t - R_t \right|$$
17:  Sort the distances in decreasing order
18:  Compute $\theta$ using (27), i.e., $\theta = d(\lfloor \eta * N \rfloor)$
19:  Use (28) to decide whether a new sample $z$ belongs to target or not
     $$Sign(\theta - d(z)) = \begin{cases} 1, & z \text{ is classified as target} \\ -1, & z \text{ is classified as outlier} \end{cases}$$
20: **end for**

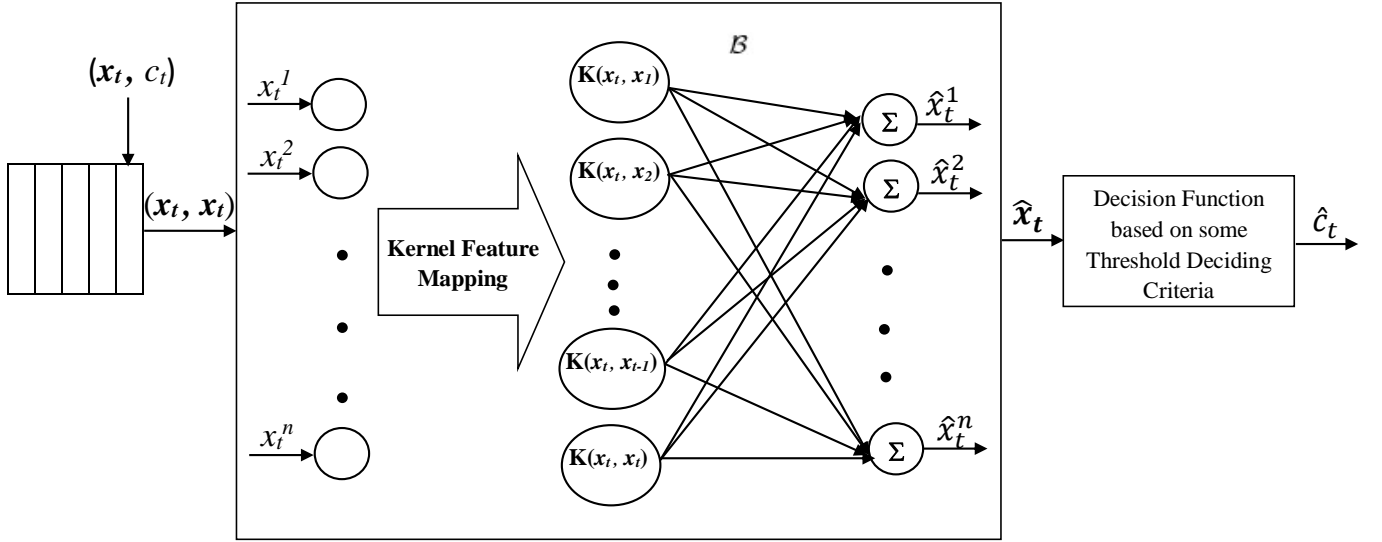

Fig. 3: Schematic Diagram of Online Sequential One-class Extreme Learning Machine: Reconstruction Based

Where $0 < \eta \leq 1$ is the fraction of rejection of training samples for deciding threshold value. $N$ is the number of training samples and $\lfloor \rfloor$ denotes floor operation. We consider 5% of rejection, i.e. $\eta = 0.05$. Now, generate a decision function to decide whether any new sample $z$ belongs to the target or outlier, where $z = [z^1, z^2, ..., z^n]$. This function is defined by the following equation:

$$Sign(\theta - d(z)) = \begin{cases} 1, & z \text{ is classified as target} \\ -1, & z \text{ is classified as outlier} \end{cases} \quad (28)$$

The above proposed approach is summarized as pseudo code in **Algorithm 1**.

### B. ORK-OCELM(R): Reconstruction Based Approach

In reconstruction framework based **ORK-OCELM** i.e. **ORK-OCELM(R)**, model is trained by only target data $X$ and approximates all data to itself. Fig. 3 shows the architecture for **ORK-OCELM(R)**. Here, the given stream of training data $X$ is defined same as discussed above in the boundary framework. Input layer that takes data for $t^{th}$ input sample is coded as $(x_t, x_t)$ because the target is identical to the input layer. Further, kernel feature mapping is employed between input and hidden layer, which is same as discussed in the above section. Output weight matrix $\mathcal{B}$ until $t^{th}$ sample is represented as $[\mathcal{B}_1, \mathcal{B}_2....\mathcal{B}_{(t-1)}, \mathcal{B}_t]$. $\widehat{X} = (\widehat{x}_1, \widehat{x}_2...\widehat{x}_t,...)$ is the set of predicted output vectors, where $\widehat{x}_t = [\widehat{x}_t^1, \widehat{x}_t^2, ..., \widehat{x}_t^n]$ is the predicted output vector for the $t^{th}$ sample. $\widehat{c} = [\widehat{c}_1, \widehat{c}_2...\widehat{c}_t, ...]$ is the predicted class vector, where $\widehat{c}_t$ is the predicted class for the $t^{th}$ sample. Formulation of **ORK-OCELM(R)** is as discussed next.

Similar to **ORK-OCELM(B)**, minimization problem here is first written for random feature mapping using hidden layer matrix $H_0$ and their weight matrix $W_0$. Later, this formulation is derived for kernel feature mapping using Representer Theorem [58].

For the initial chunk $\{X_0, X_0\}$ of size $N_0$, the objective is to minimize the output weight $W_0$ (output weight matrix for the random feature mapping case) and the reconstruction error $E_0$ between the expected ($X_0$) and the predicted values ($W_0^T H_0$). Minimization problem for the case of random feature mapping can be written as follows:

$$\text{Minimize}: L = \frac{1}{2}\|W_0\|^2 + \lambda\frac{1}{2}\|E_0\|^2 \\ \text{Subject to}: W_0^T H_0 = X_0 - E_0 \quad (29)$$

Here, $\lambda$ is used as a regularization parameter.

Next, Representer Theorem [58] is exploited, which describes the weight matrix $W_0$ as a linear combination of the training data representation in ELM space ($H_0$) and a reconstruction matrix $\mathcal{B}_0$. That is, the output weight matrix for the kernel feature mapping case is given as follows:

$$W_0 = H_0 \mathcal{B}_0 \quad (30)$$

It is to be noted that $w_0$ and $\beta_0$ are weight vectors due to single output node architecture in **ORK-OCELM(B)**, however, $W_0$ and $\mathcal{B}_0$ are weight matrices due to multi-output node architecture in **ORK-OCELM(R)**.

Thus, the minimization problem in (29) can be reformulated as follows:

$$\text{Minimize}: L = \frac{1}{2}\mathcal{B}_0^T H_0^T H_0 \mathcal{B}_0 + \lambda\frac{1}{2}\|E_0\|^2 \\ \text{Subject to}: \mathcal{B}_0^T H_0^T H_0 = X_0 - E_0 \quad (31)$$

Now, by substituting kernel matrix $\Phi_0 = \phi(X_0) = (H_0)^T H_0 = K(X_0, X_0)$ in the above equation, the following minimization problem is obtained:

$$\text{Minimize}: L = \frac{1}{2}\mathcal{B}_0^T \Phi_0 \mathcal{B}_0 + \lambda\frac{1}{2}\|E_0\|^2 \\ \text{Subject to}: \mathcal{B}_0^T \Phi_0 = X_0 - E_0 \quad (32)$$

By using Lagrangian relaxation [59], (32) can be written as dual optimization problem:

$$\text{Minimize}: L = \frac{1}{2}\mathcal{B}_0^T \Phi_0 \mathcal{B}_0 + \lambda \frac{1}{2} \|E_0\|^2 \\ -\alpha(\mathcal{B}_0^T \Phi_0 - X_0 + E_0) \quad (33)$$

where $\alpha$ is the Lagrange multiplier, employed to combine the constraint with the objective function. By using Karush-Kuhn-Tucker (KKT) theorem [60], take partial derivatives of (33) with respect to all variables $\mathcal{B}_0, E_0$ and $\alpha$. That is,

$$\begin{aligned} \frac{\partial L}{\partial \mathcal{B}_0} &= 0 \Rightarrow \mathcal{B}_0 = \alpha \\ \frac{\partial L}{\partial E_0} &= 0 \Rightarrow \lambda E_0 = \alpha \\ \frac{\partial L}{\partial \alpha} &= 0 \Rightarrow \mathcal{B}_0^T \Phi_0 - X_0 + E_0 = 0 \end{aligned} \quad (34)$$

Following weight is obtained from solving equation in (34):

$$\mathcal{B}_0 = P_0 X_0, \text{ where } P_0 = \left(\Phi_0 + \frac{1}{\lambda}I\right)^{-1} \quad (35)$$

Finally, weight matrix for initial samples is obtained, which needs to be updated as follows:

Initially, $\Phi$ and $P$ are calculated as per (35). Both are updated continuously for any upcoming new samples $X^v$ as per (11)-(19). After processing the training dataset $X_{curr}^{tr}$ in the current sliding window, output function for any set of k samples $X_k = \{x_1, x_2....x_k\}$ can be written as follows:

$$f(X_k) = \begin{bmatrix} K(X_{curr}^{tr}, x_k) \\ \vdots \\ K(X_{curr}^{tr}, x_1) \end{bmatrix}^T P X_{curr}^{tr} \quad (36)$$

where $K(X_{curr}^{tr}, x_i)$ denotes kernel vector for the $i^{th}$ sample $x_i$.

Afterwards, calculate error and decide whether any sample belongs to target class or outlier class as per pseudo code discussed in **Algorithm 2**.

## III. PERFORMANCE EVALUATION

In this section, the performance of the classifiers is tested on both types of the environments, i.e. stationary and non-stationary. Two synthetic and six real time datasets are used for the stationary environment (see Table II) and, sixteen synthetic and four real time datasets are used for the non-stationary environment (see Table IV). All experiments are executed on MATLAB 2014b in Windows 7 (64 bit) environment with 64 GB RAM, 3.00 GHz Intel Xeon processor. For implementing the existing classifiers, two toolboxes are mainly used (developed by Tax [63] and Gautam et al. [27]). All existing and proposed classifiers are implemented and tested in the same environment. Their parameters are tuned using consistency based model selection approach, which is discussed below.

---

**Algorithm 2** *ORK-OCELM(R)*:Reconstruction Based

**Input:** Training set $X = (x_1, c_1), (x_2, c_2), ..., (x_{N_0}, c_{N_0}), ..., (x_t, c_t), ...$

**Output:** Whether each sample belongs to Target class or Outlier class

**Initialization Phase**

1: Pass initial set of samples $X_0$ to the classifier as: $\{(x_1, x_1), (x_2, x_2), ..., (x_{N_0}, x_{N_0})\}$
 // For first chunk of $N_0$ samples, follow the steps as below.
2: Employ kernel feature mapping: $\Phi_0 = \phi(X_0)$.
3: Output Weight $\mathcal{B}_0$ for $(X_0, X_0)$:
$$\mathcal{B}_0 \leftarrow P_0 X_0$$
$$P_0 \leftarrow \Phi_0^{-1}$$

**End of Initialization Phase**
 // For second chunk onwards, following steps are required
$$\Phi \leftarrow \Phi_0$$
$$P \leftarrow P_0$$

4: **for** $i = 1$ to last chunk of data in $X$ **do**
5:  Set size of chunk at the current stage as $s$
6:  Update the final kernel matrix $\Phi$ and its inverse $P$ in two steps as given below:
  **Step 1: Forgetting Phase**
7:  Remove the impact of $s$ old samples from the current inverse $P$ using (24), i.e.,
$$P_{new} = R_{22}^{-1} = Ri_{22} - Fi_{12}^T Fi_{11}^{-1} Fi_{12}$$
8:  Update the kernel matrix $\Phi$ also by removing those rows and columns which were generated due to the old $s$ samples.
  **Step 2: Retraining Phase**
9:  Update the kernel matrix $\Phi$ as per (11), i.e.,
$$\Phi = \begin{bmatrix} \Phi_u & \Phi_{u,v} \\ (\Phi_{u,v})^T & \Phi_v \end{bmatrix}$$
10:  Calculate $\Phi_v$ for $i^{th}$ chunk by using (12)
11:  Calculate $\Phi_{u,v}$ for $i^{th}$ chunk by using (13)
12:  Compute the inverse of updated kernel matrix $\Phi$, i.e., $P = \Phi^{-1}$ using block inverse as discussed in (15)-(19)
13:  $b = b + s$
14:  Update the Output Weight using the value of $X_{curr}^{tr}$ and updated $P$ as $\mathcal{B} = P X_{curr}^{tr}$
15:  Compute the predicted value by using output function $f(X_k)$, which is defined in (36)
16:  Calculate sum of square error $(d)$ between predicted $(\hat{x}_t)$ and actual value $(x_t)$ of $t^{th}$ training sample:
$$d(x_t) = \sum_{j=1}^{n}(x_t^j - \hat{x}_t^j)^2 \quad (37)$$
17:  Sort the distances in decreasing order
18:  Compute $\theta$ using (27), i.e., $\theta = d(\lfloor \eta * N \rfloor)$
19:  Use (28) to decide whether a new sample $z$ belongs to target or not
$$Sign(\theta - d(z)) = \begin{cases} 1, & z \text{ is classified as target} \\ -1, & z \text{ is classified as outlier} \end{cases}$$
20: **end for**

## A. *Parameter Selection Using Consistency based Model Selection for ELM based One-class Classifier*

Parameter selection is always a crucial task in the case of one-class classification. Among various optimal parameter selection methods [64] [65], consistency based model selection [66] is the only method, which is suitable for any type of kernels. Consistency based model selection determines a threshold error ($E_{thr}$) by using (38) given below, and executes the classifier with different combination of parameters until it yields less error than $E_{thr}$.

$$E_{thr} = (M * \eta + \sigma_{thr} * \sqrt{(\eta * (1-\eta) * M)})/M$$
$$= \eta + \sigma_{thr} * \sqrt{(\eta * (1-\eta)/M)} \quad (38)$$

where

(i) $M$ denotes number of samples in the validation set.
(ii) $M = (N/f)$, where $N$ denotes number of training samples and $f$ is number of folds.
(iii) $\sigma_{thr}$ is the threshold required for determining decision boundary during model selection.
(iv) $\eta$ denotes the fraction of the samples rejected from the dataset, and lies between 0 and 1.
(v) $(M * \eta)$ is the expected number of rejected samples.
(vi) $\eta * (1-\eta) * M$ is the variance
(vii) $(M * \eta + \sigma_{thr} * \sqrt{(\eta * (1-\eta) * M)})$ is the upper limit to the number of rejected target objects.

For the proposed one-class classifiers, every possible combination of $\sigma$ and $\lambda$ are employed to obtain the most complex classifier as long as classifier is consistent. The range of $\lambda$ considered is $[10^{-8}, 10^{-7}, ..., 10^{7}, 10^{8}]$ and the range of $\sigma$ considered is twenty values between minimum and maximum pairwise distance among training samples of the dataset.

TABLE II: Dataset Description for stationary Datasets

| Real world stationary Datasets | | | | | |
|---|---|---|---|---|---|
| | #Feat. | Drift | #Target | #Outlier | #Records |
| Breast Cancer [67] | 9 | No | 241 | 458 | 699 |
| Diabetes [67] | 8 | No | 500 | 268 | 768 |
| Ecoli [67] | 7 | No | 52 | 284 | 336 |
| Liver [67] | 6 | No | 145 | 200 | 345 |
| Sonar [67] | 60 | No | 111 | 97 | 208 |
| Spectf [67] | 44 | No | 95 | 254 | 349 |

## B. *Performance Comparison on Stationary Datasets*

As one-class classifiers are also called as data descriptors, and hence, the proposed methods are first tested on stationary synthetic datasets to verify their data describing (boundary creation) ability. The motive to test on the stationary datasets is to show that online classifiers perform at least as good as their corresponding batch learning based classifiers. As, **ORK-OCELM(B)**, **ORK-OCELM(R)** are the online version of OCKELM [24] and AAKELM [27], respectively, we compare against these two. It can be easily visualized in Fig. 4 that the proposed online classifiers have created similar boundary like their corresponding offline one-class classifiers. Boundary creation for ring dataset is challenging task for one-class classifiers as these data can be covered by a single circle as well as by two concentric circles. However, two concentric circles are a more appropriate boundary, which can be seen in Fig. 4a to 4d. It can be observed in Fig. 4a to 4h that the boundary based classifiers create smoother boundary compared to the reconstruction based classifier for both the synthetic datasets. For real world datasets, each experiment is repeated over twenty runs and then the average of the Area Under Curve ($AUC = 1/2$(Sensitivity + Specificity)) is calculated (see Table III). Results are reported in terms of $AUC$ because different classes of these datasets are slightly imbalanced and $AUC$ provides a better insight compared to Accuracy ($Acc$) in such cases. This is because $AUC$ considers sensitivity and specificity with equal weight. It can be observed from Table III that the proposed online one-class classifiers yield similar or better performance compared to the offline one-class classifiers for all the datasets.

TABLE III: Performance Comparison on Stationary Datasets in term of average AUC over 20 Runs

| | Breast Cancer | Diabetes | Ecoli | Liver | Sonar | Spectf |
|---|---|---|---|---|---|---|
| knndd | **95.36** | 59.82 | **89.87** | 51.73 | 45.72 | 74.98 |
| svdd | 83.17 | 50 | 72.52 | 51.96 | 56.08 | 68.2 |
| kmeans_dd | 93.59 | 61.65 | 88.84 | 51.83 | 58.37 | 76.34 |
| parzen_dd | 94.37 | 59.94 | 89.65 | 50.97 | 50 | 68.88 |
| nparzen_dd | 92.51 | **64.36** | 79.47 | 51.53 | 45.23 | 60.56 |
| pca_dd | 88.19 | 56.25 | 62.59 | 52.67 | 59.45 | 73.4 |
| mpm_dd | 77.89 | 54.63 | 76.04 | 54.51 | 50 | 68.56 |
| incsvdd | 94.36 | 57.15 | 76.39 | 50.79 | 61.84 | 75.87 |
| som_dd | 90.29 | 64.26 | 56.47 | 53.41 | 63.59 | 69.4 |
| autoenc_dd | 93.05 | 58.46 | 80.39 | 53.52 | 47.41 | 76.37 |
| gauss_dd | 95.11 | 57.12 | 88.68 | 51.84 | 50 | 67.77 |
| OCKELM | 93.68 | 64.07 | 86.04 | 54.16 | 64.16 | 77.48 |
| AAKELM | 93.94 | 61.33 | 63.25 | 52.24 | 46.84 | 75.87 |
| Proposed Methods | | | | | | |
| ORK-OCELM(B) | 95.22 | 64.11 | 88.12 | **56.72** | **67.74** | **78.95** |
| ORK-OCELM(R) | 93.63 | 61.24 | 62.62 | 52.12 | 47.6 | 75.26 |

TABLE IV: Dataset Description for Non-stationary Datasets

| Synthetic Non-stationary Datasets | | | | | |
|---|---|---|---|---|---|
| Dataset Name | #Feat. | Drift | #Target | #Outlier | #Records |
| 1CDT [68] | 2 | 400 | 8000 | 8000 | 16,000 |
| 1CHT [68] | 2 | 400 | 8000 | 8000 | 16,000 |
| 4CE1CF [68] | 2 | 750 | 34650 | 138600 | 173,250 |
| 4CR [68] | 2 | 400 | 36100 | 108300 | 144,400 |
| GEARS-2C-2D [69] | 2 | 2,000 | 100000 | 100000 | 200,000 |
| 4CRE-V1 [68] | 2 | 1,000 | 31250 | 93750 | 125,000 |
| 4CRE-V2 [68] | 2 | 1,000 | 45750 | 137250 | 183,000 |
| 5CVT [68] | 2 | 1,000 | 8000 | 16000 | 24,000 |
| 2CDT [68] | 2 | 400 | 8000 | 8000 | 16,000 |
| 2CHT [68] | 2 | 400 | 8000 | 8000 | 16,000 |
| 1CSurr [68] | 2 | 600 | 20200 | 35083 | 55,283 |
| UG-2C-2D [69] | 2 | 1,000 | 50000 | 50000 | 100,000 |
| UG-2C-3D [69] | 3 | 2,000 | 100000 | 100000 | 200,000 |
| UG-2C-5D [69] | 5 | 2,000 | 100000 | 100000 | 200,000 |
| MG-2C-2D [69] | 2 | 2,000 | 100000 | 100000 | 200,000 |
| FG-2C-2D [70] | 2 | 2,000 | 150000 | 50000 | 200,000 |

| Real Non-stationary Datasets | | | | | |
|---|---|---|---|---|---|
| | #Feat. | Drift | #Target | #Outlier | #Records |
| Electricity [71] | 8 | Real | 19237 | 26075 | 45,312 |
| Poker-hand [67] | 10 | Real | 415526 | 413675 | 829,201 |
| Keystroke [72] | 10 | Real | 400 | 1200 | 1600 |
| Abalone [67] | 10 | Real | 2770 | 1407 | 4177 |

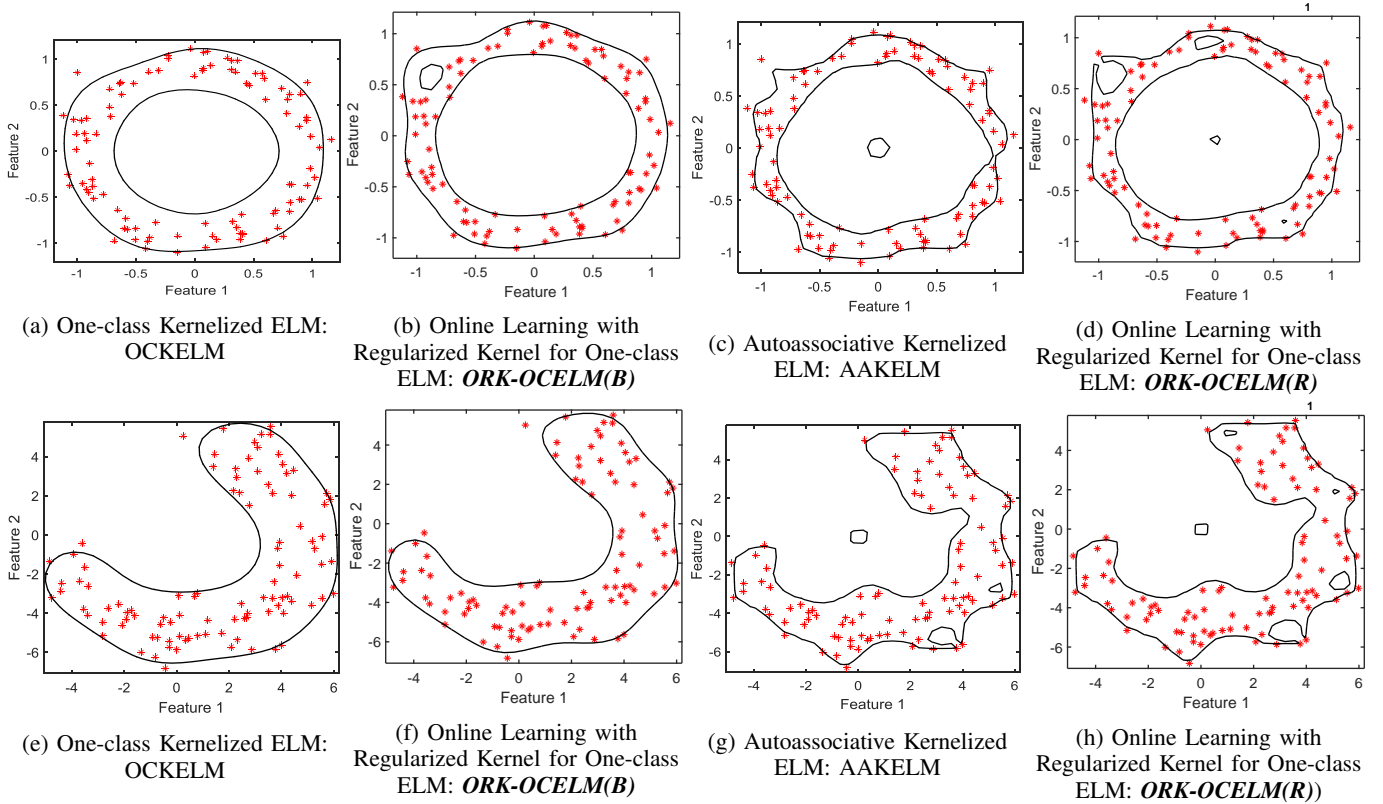

Fig. 4: Performance of the Proposed Online Classifiers on Synthetic dataset:Boundary and Reconstruction based approaches

### C. Performance Comparison on Non-stationary Datasets

The proposed online one-class classifiers are tested on non-stationary datasets to verify two things; first, adaptiveness and second, large streaming datasets handling capability with limited memory. All non-stationary synthetic datasets have different types of drift associated with them like periodic drift, non-periodic drift etc. Drift occurrence of each dataset is briefly described in Table IV and also visualized on the web page (https://goo.gl/j2wQw4). The synthetic datasets used in this paper are taken from ( [68]–[70] and [72]). All synthetic and real datasets were originally proposed for binary or multi-class classification in the non-stationary environment. We modify these datasets for the OCC task by assuming one-class as a normal class and the remaining all classes as an outlier class.

Classifiers are experimented in two modes viz., static (i.e. offline classifiers) and sliding window (i.e. online classifiers). One-class classifiers in static mode are named as OCKELM(S) and AAKELM(S). Here, 'S' stands for static mode. In static mode, one-class classifiers are just trained on initially available samples and then tested on all remaining samples like batch learning. ***ORK-OCELM(B)***, ***ORK-OCELM(R)***, incSVDD [22] and OKPCA [23] are tested in the sliding window mode. Here, the classifiers adapt to the environment as per the upcoming new samples in sliding window and forget the old samples. Except for the large real world datasets, the sliding window and the block size are fixed as 150 and 50, respectively (see the last two columns of Table V and VI).

The accuracy of these datasets is plotted in Fig. 5 and 6. These plots are created by dividing the dataset in hundred batches and the mean accuracy achieved by all the batches in 100 steps is plotted. Here, the y-axis and the x-axis denote accuracy and steps, respectively for all of the plots in Fig. 5 and 6.

Next, we discuss results for different datasets based upon the nature of the drift.

(a) **Drift only in the outlier class:** The three datasets that have this behavior are 1CDT, 1CHT and 4CE1CF. It can be observed from Table V and Fig. 5a, 5b, 5c that one-class classifiers in static mode yield similar or comparative results as a sliding window mode for these datasets. This is because the normal class in these datasets are not changing their distribution over time, only the outlier class changes its distribution. However, the same can't be stated for the remaining synthetic datasets as the distribution of both classes is changing in the remaining datasets. OKPCA performs better for 1CDT and 1CHT datasets, however, ***ORK-OCELM(B)*** performs better for 4CE1CF datasets compared to the other online one-class classifiers, which are present in Table V.

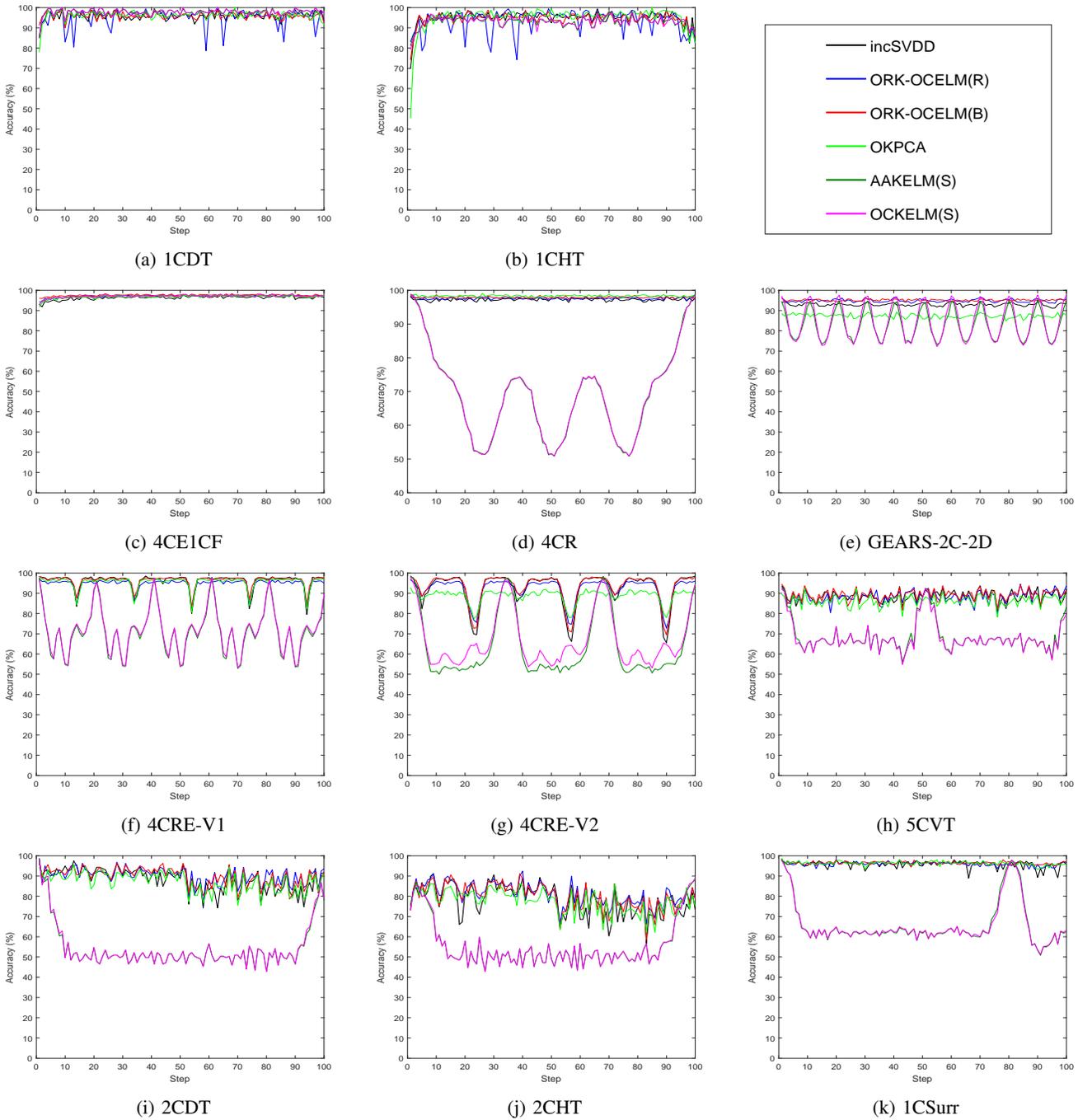

Fig. 5: Performance of one-class classifiers on all datasets in 100 steps. This figure is continued to Fig. 6

(b) **Periodic drift in outlier and non-periodic drift in the normal class:** The dataset which has this behavior is 4CR. For 4CR dataset, the normal class rotates on a circular path and three different distributions of outlier also rotate on the circular path but they create a periodic drift collectively. Periodic drift can also be visualized by observing the repeating nature of the generated performance plot for 4CR dataset of the classifiers in the static mode (see Fig. 5d). As expected static mode doesn't perform well for this dataset and the proposed classifiers yield better accuracy compared to all except OKPCA.

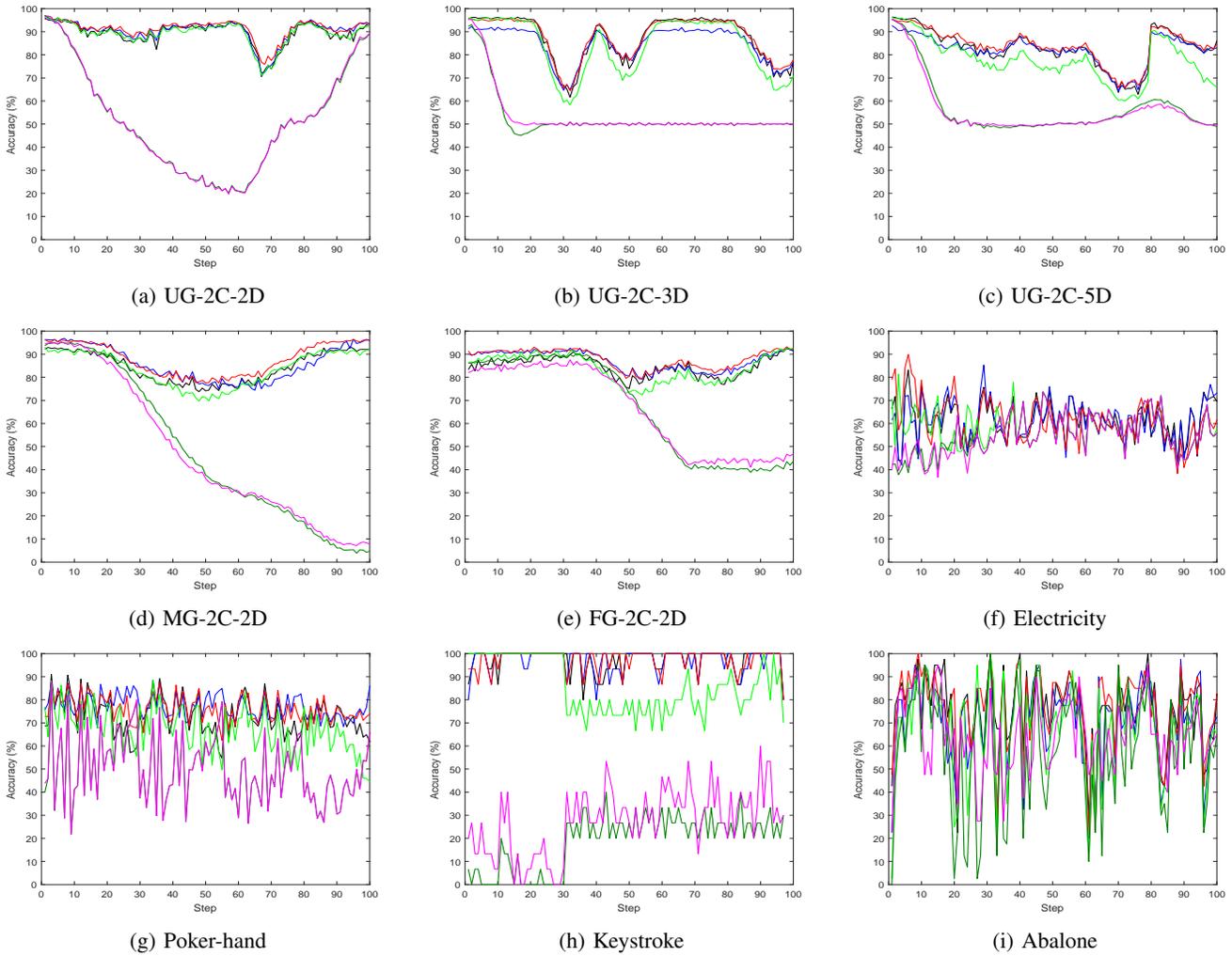

Fig. 6: Continuation of Fig. 5

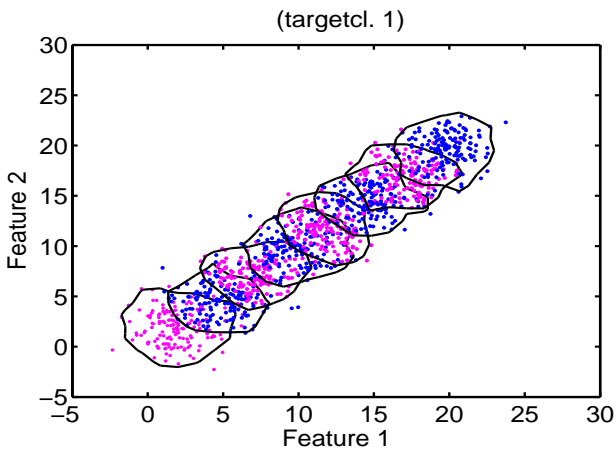

Fig. 7: Adaption of ***ORK-OCELM(B)*** on 2CDT dataset

(c) **Periodic drift in both the normal and the outlier class:** The four datasets that have this behavior are Gears-2C-2D, 4CRE-V1, 4CRE-V2 and 5CVT. For these datasets, classifiers in static mode yield inferior performance compared to sliding window mode classifiers due to the periodic drift occurrence in the normal as well as the outlier class, which can also be noticed in the Fig. 5e, 5f, 5g and 5h (the performance dips down at regular periodic interval for static mode). However, the proposed and existing online classifiers smoothly adapt to this periodic drift. The adaptation of ***ORK-OCELM(B)*** in the non-stationary environment has been visualized for Gears-2C-2D dataset along with some other datasets on the web page(https://goo.gl/8bvkps). As the Gears-2C-2D dataset has a smaller drift, and therefore, classifiers in static mode perform better for this dataset compared to other three synthetic datasets. Boundary based one-class classifier, i.e. ***ORK-OCELM(B)*** outperforms all the present classifiers in Table V. Specifically, ***ORK-OCELM(B)*** outperforms the OKPCA by significant margin, i.e. 4.51%, 2.83%, 7.74% and 0.87%, for 4CRE-V2, 5CVT, Gears-2C-2D and 4CRE-V1, respectively.

(d) **Non-periodic drift in both the normal and the outlier class:** The two datasets that have this behavior are 2CDT and 2CHT. Drift continuously increases diagonally and horizontally for these datasets, respectively. Adap-tion of **ORK-OCELM(B)** on 2CDT dataset is visualized in the Fig. 7. Here, all plotted samples belong to the normal class but at distinct timestamps. Two colors are used to discriminate the normal samples at two consecutive timestamps. Reconstruction based classifier, i.e. **ORK-OCELM(R)** performs best among all the classifiers and our proposed classifiers outperform incSVDD and OKPCA significantly by more than 2% and 3% margin for 2CDT and 2CHT datasets, respectively. Among all sixteen synthetic datasets, proposed and existing one-class classifiers achieve the least accuracy for 2CHT dataset, i.e. when horizontal non-periodic drift occurs.

(e) **Drift created by surrounding the normal class by outlier:** In 1CSurr dataset, normal and outlier, both classes are drifting and normal data is compactly surrounded by outliers in a non-periodic way. **ORK-OCELM(B)** performs best among all, however, **ORK-OCELM(R)** yields better accuracy than incSVDD but inferior to OKPCA.

Next, we discuss results for different datasets based upon the occurrence of drift with Gaussian distribution.

(a) **Drift as per Unimodal Gaussian distribution of the normal and the outlier class:** Normal and the outlier classes are generated by unimodal Gaussian distribution for three datasets viz., UG-2C-2D, UG-2C-3D and UG-2C-5D, of two, three and five dimensions, respectively. Distribution of UG-2C-2D and the adaptive boundary created by the **ORK-OCELM(B)** on this dataset for different timestamps are visualized in the Fig. 8. Distribution of these datasets are also visualized on this web page (https://goo.gl/j2wQw4) and adaptiveness of **ORK-OCELM(B)** on UG-2C-2D is visualized on this web page (https://goo.gl/8bvkps). **ORK-OCELM(B)** performs best among all and outperforms incSVDD and OKPCA for all three datasets. Overall, **ORK-OCELM(R)** performs well on the two dimensional dataset, however, for three and five dimensional datasets, i.e. UG-2C-3D and UG-2C-5D, it still performs better than OKPCA but inferior to incSVDD.

(b) **Drift as per Multi-modal Gaussian distribution of the normal and the outlier class:** The normal and the outlier classes are generated by multi-modal Gaussian distribution for two datasets viz., MG-2C-2D and FG-2C-2D. Distribution of MG-2C-2D and the adaptive boundary created by the **ORK-OCELM(B)** on this dataset for different timestamps are visualized in the Fig. 9. In the MG-2C-2D dataset, normal and outlier both show multi-modal behavior alternatively. However in FG-2C-2D dataset, multi-modality is only present in the normal class, which is represented by four distinct Gaussian distribution (The outlier class has single Gaussian distribution). More detailed visualization of both datasets as well as behavior of **ORK-OCELM(B)** on these datasets are available on these web pages(https://goo.gl/j2wQw4 and https://goo.gl/8bvkps). **ORK-OCELM(B)** outperforms incSVDD by more than 3% and OKPCA by nearly 4% of margin. After 2CHT dataset, classifiers yield the least accuracy on MG-2C-2D dataset due to the presence of multimodality in this dataset.

### D. Drift in non-stationary real world datasets

Four real world datasets are employed for the performance analysis. Electricity dataset is a well know dataset for streaming data analysis. It is collected from markets where the price varies as per demand and supply. The main task here is to identify the change of the price relative to a moving average of the last 24 hours. Proposed classifiers clearly outperform static mode based classifiers as well as other online classifiers available in Table VI. In Poker-hand dataset, original dataset contains ten classes where one class belongs to non-poker hand and the remaining classes belong to nine types of poker hands. We club these nine poker hands and treat them as a single class. The main task here is to identify the poker hand after training the one-class classifier on non-poker hand. As seen in Table VI and Fig. 6g, **ORK-OCELM(R)** performs best among all the classifiers for this dataset and both of the proposed classifiers outperform incSVDD and OKPCA by the significant margin of more than 3% and 10% respectively.

The third dataset is a Keystroke dynamics dataset, where user verification is based on the typing rhythm of the user instead of the traditional way (login id and password) without any extra cost of hardware. We need to update the user profile regularly as typing rhythm of a user evolves over time, and therefore, the distribution also changes and drift occurs. Classifier is trained on the data of one user and then the trained model is used for the verification purpose of that user. As seen in Table VI and Fig. 6h, **ORK-OCELM(B)** yields the best accuracy among all and significantly outperforms OKPCA, however, yields slightly better accuracy compared to incSVDD. The fourth dataset, i.e. Abalone dataset, contains originally twenty-nine classes, which are converted for the OCC task by considering classes 9-29 as the normal class and class 1-8 as the outlier class. As seen in Table VI and Fig. 6i, **ORK-OCELM(B)** performs best among all for this dataset, however, **ORK-OCELM(R)** performs better than OKPCA but inferior to incSVDD.

### E. Efficiency Analysis

Time efficiency is the main concern in stream processing for various applications. The proposed classifiers are very simple and computationally more efficient compared to other classifiers. Since the presented online classifiers are based on ELM, and therefore, they inherit the fast learning property of ELM also. For verifying this fact in an unbiased manner, existing classifiers are implemented and tested in the same environment as the proposed classifiers. We have reported time consumed by the classifiers in three parts viz., training, forgetting and testing time in in Table VII. We are reporting forgetting time explicitly in Table VII since it is crucial in online learning.

The reason behind computational efficiency of the proposed algorithms can be understood by the analysis of (18) and (19). Recall, the complexity of matrix inversion and matrix multiplication is $\mathcal{O}(n^3)$. However, this complexity is further reduced in three ways. First, if any multiplication algorithm

TABLE V: Accuracy of one-class classifiers over 16 synthetic Non-stationary datasets

|  | **ORK-OCELM(B)** | OCKELM(S) | **ORK-OCELM(R)** | AAKELM(S) | incSVDD | OKPCA | Chunk Size | Sliding Window size |
|---|---|---|---|---|---|---|---|---|
| **1CDT** | 95.96 | 97.73 | 95.30 | **97.79** | 95.85 | **96.22** | 50 | 150 |
| **2CDT** | 89.84 | 54.38 | **90.22** | 54.24 | 88.17 | 87.34 | 50 | 150 |
| **1CHT** | 94.56 | 93.16 | 93.60 | 93.11 | 94.89 | **94.97** | 50 | 150 |
| **2CHT** | 80.04 | 55.64 | **81.08** | 55.61 | 78.06 | 77.50 | 50 | 150 |
| **4CR** | 97.95 | 69.23 | 97.58 | 69.17 | 97.26 | **98.34** | 50 | 150 |
| **4CRE-V1** | **96.27** | 71.99 | 94.74 | 71.44 | 96.10 | 95.40 | 50 | 150 |
| **4CRE-V2** | **93.19** | 67.11 | 92.26 | 63.76 | 92.25 | 88.68 | 50 | 150 |
| **5CVT** | **89.22** | 68.26 | 88.41 | 68.68 | 88.55 | 86.38 | 50 | 150 |
| **1CSurr** | **96.46** | 66.00 | 95.99 | 66.06 | 95.70 | 96.27 | 50 | 150 |
| **4CE1CF** | **97.58** | 97.09 | 97.12 | 97.03 | 96.32 | 96.75 | 50 | 150 |
| **FG-2C-2D** | **88.04** | 65.78 | 87.21 | 66.49 | 84.72 | 84.09 | 50 | 150 |
| **UG-2C-2D** | **91.22** | 51.16 | 90.42 | 51.13 | 89.54 | 89.36 | 50 | 150 |
| **UG-2C-3D** | **87.77** | 53.57 | 84.84 | 52.99 | 87.45 | 84.11 | 50 | 150 |
| **UG-2C-5D** | **84.12** | 56.09 | 82.14 | 56.74 | 83.10 | 77.79 | 50 | 150 |
| **MG-2C-2D** | **87.99** | 46.80 | 86.06 | 47.56 | 84.51 | 83.79 | 50 | 150 |
| **GEARS-2C-2D** | **95.12** | 84.59 | 94.89 | 83.22 | 92.99 | 87.37 | 50 | 150 |

TABLE VI: Accuracy of one-class classifiers over real world Non-stationary datasets

|  | **ORK-OCELM(B)** | OCKELM(S) | **ORK-OCELM(R)** | AAKELM(S) | incSVDD | OKPCA | Chunk Size | Sliding Window size |
|---|---|---|---|---|---|---|---|---|
| **Electricity** | **61.64** | 55.03 | 61.49 | 55.14 | 61.25 | 58.67 | 200 | 2500 |
| **Poker-hand** | 76.12 | 49.96 | **77.18** | 49.90 | 73.29 | 66.95 | 200 | 2500 |
| **Keystroke** | **97.38** | 27.38 | 96.97 | 18.83 | 97.31 | 85.86 | 50 | 150 |
| **Abalone** | **76.73** | 64.27 | 70.72 | 57.66 | 75.32 | 66.77 | 50 | 150 |

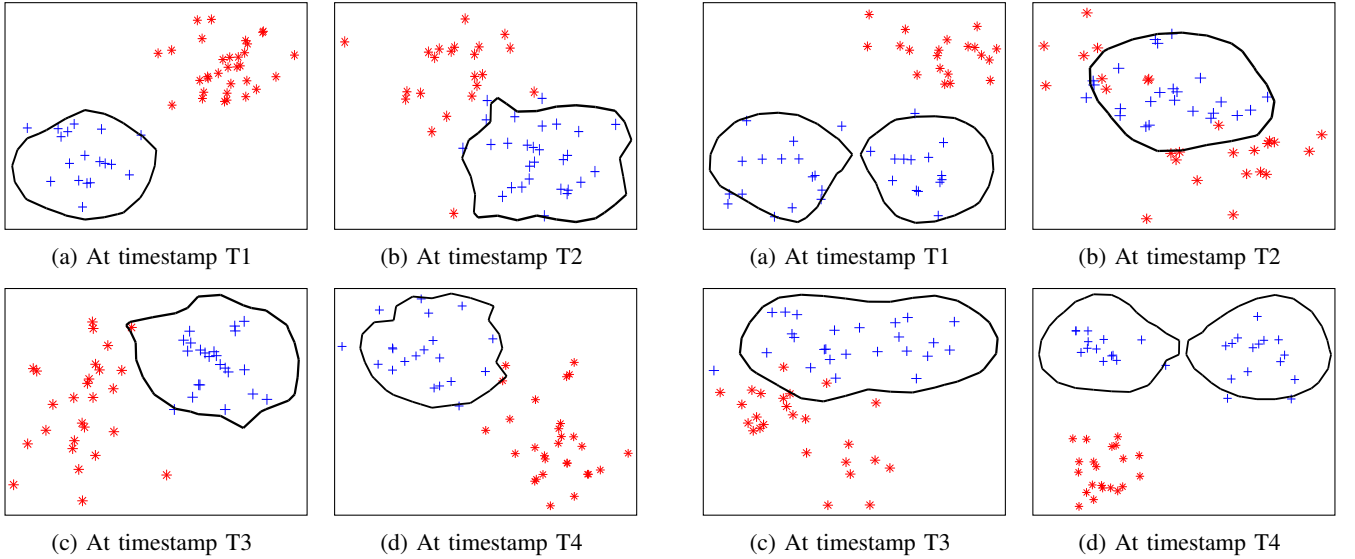

Fig. 8: Adaption on UG-2C-2D dataset

Fig. 9: Adaption on MG-2C-2D dataset

(like Strassen matrix multiplication) of complexity $\mathcal{O}(n^{2+\epsilon})$ is used, then the complexity of block matrix inversion is further reduced to $\mathcal{O}(n^{2.807})$ [73]. Second, Woodbury formula [62] is used to solve (18) as shown in (19). After using Woodbury formula [62] in (19), the inverse of two matrices, i.e. $\mathbf{\Phi_u}$ and $\mathbf{\Phi_v}$ are required to be computed. Here, $\mathbf{\Phi_u}$ is the larger matrix of the size of the previous sliding window and $\mathbf{\Phi_v}$ is the smaller matrix of the size of newly added samples for training. As per **Algorithm 1** and **Algorithm 2**, $\Phi_u^{-1}$ is already computed during the calculations for previous sliding window. Therefore, there is no need to compute $\Phi_u^{-1}$ for the current window and only the inverse of the smaller size of the matrix $\Phi_v$ needs to be calculated, which is the third improvement. Similarly, the forgetting mechanism in (24) is also employed by available inverses instead of calculating inverses explicitly. Hence, overall computational time cost is further reduced to $\mathcal{O}(n^2)$. Moreover, Time is also saved during kernel matrix calculation as it reuses the previously computed kernel matrix for the data from previous sliding window. Next, we discuss storage cost.

TABLE VII: Time (sec.) consumes during training, forgetting and testing for Sliding Window size=150 and chunk size = 50

|  | *ORK-OCELM(B)* | | | *ORK-OCELM(R)* | | | incSVDD | | |
|---|---|---|---|---|---|---|---|---|---|
|  | Training | Forgetting | Testing | Training | Forgetting | Testing | Training | Forgetting | Testing |
| **1CDT** | 1.47 | 0.64 | 128.55 | 1.40 | 0.52 | 126.51 | 9.05 | 8.31 | 155.24 |
| **2CDT** | 1.42 | 0.64 | 127.64 | 1.48 | 0.59 | 120.43 | 8.40 | 7.83 | 142.28 |
| **1CHT** | 1.41 | 0.60 | 126.13 | 1.35 | 0.51 | 124.25 | 9.06 | 8.35 | 152.65 |
| **2CHT** | 1.39 | 0.64 | 126.64 | 1.44 | 0.59 | 119.50 | 8.50 | 7.83 | 141.15 |
| **4CR** | 11.48 | 5.27 | 1149.44 | 12.27 | 5.15 | 1091.52 | 37.31 | 34.71 | 1292.02 |
| **4CRE-V1** | 9.06 | 3.93 | 873.20 | 9.26 | 3.66 | 858.66 | 32.04 | 30.13 | 1045.92 |
| **4CRE-V2** | 13.87 | 6.17 | 1371.48 | 14.34 | 5.94 | 1323.65 | 46.88 | 43.76 | 1584.57 |
| **5CVT** | 1.83 | 0.81 | 170.08 | 1.80 | 0.72 | 165.89 | 8.34 | 7.88 | 201.62 |
| **1CSurr** | 4.46 | 1.90 | 403.17 | 4.26 | 1.66 | 394.50 | 20.90 | 19.34 | 480.99 |
| **4CE1CF** | 11.62 | 5.06 | 1151.00 | 11.51 | 4.57 | 1146.30 | 134.45 | 112.72 | 1542.84 |
| **FG-2C-2D** | 16.58 | 6.64 | 1408.04 | 16.78 | 6.25 | 1385.00 | 141.94 | 129.59 | 1685.98 |
| **UG-2C-2D** | 8.05 | 3.40 | 720.91 | 8.03 | 3.15 | 707.01 | 54.74 | 52.86 | 856.55 |
| **UG-2C-3D** | 15.71 | 6.48 | 1415.80 | 15.85 | 6.25 | 1391.73 | 94.47 | 86.05 | 1693.65 |
| **UG-2C-5D** | 15.70 | 6.41 | 1404.62 | 15.36 | 5.93 | 1383.24 | 96.68 | 89.43 | 1688.55 |
| **MG-2C-2D** | 15.69 | 6.57 | 1416.33 | 15.24 | 5.86 | 1389.52 | 93.32 | 83.85 | 1701.95 |
| **GEARS-2C-2D** | 15.36 | 6.39 | 1391.69 | 14.97 | 5.77 | 1371.26 | 93.75 | 84.74 | 1682.46 |
| **ELEC** | 3.75 | 1.62 | 341.49 | 3.71 | 1.46 | 334.26 | 19.40 | 17.19 | 404.73 |
| **Poker-hand** | 765.35 | 329.13 | 6119.51 | 768.97 | 349.61 | 6045.79 | 789.16 | 798.12 | 7025.56 |
| **Keystroke** | 0.13 | 0.08 | 10.66 | 0.09 | 0.04 | 10.35 | 0.43 | 0.28 | 12.55 |
| **Abalone** | 0.40 | 0.25 | 32.81 | 0.40 | 0.16 | 30.96 | 2.79 | 2.47 | 36.47 |

As presented classifiers are based on online learning, and hence, training and testing can be performed on the fly. It means that there is no need to store any old obsolete training samples after once it has been processed. Proposed classifiers simply forget the old samples and adapt to the incoming samples. The number of training samples that need to be learned by the classifiers in one batch can be fixed using the size of the sliding window (as per memory constraint of the system). Therefore, the proposed classifiers are trained with less memory requirement.

## IV. CONCLUSION

This paper has presented online learning with the regularized kernel for one-class classification. Here, regularization parameter with kernel helps the classifier to achieve good generalization capability. Two methods viz., ***ORK-OCELM(B)*** and ***ORK-OCELM(R)***, have been developed so far in the paper, which are based on boundary and reconstruction frameworks. A forgetting mechanism is also embedded with these classifiers to remove the impact of obsolete old samples, which helps in smooth adaptation to the non-stationary environment. Consistency based model selection has been employed as a model selection criteria for all the existing and proposed one-class classifiers.

The proposed classifiers can handle data on the fly in an online and efficient manner for both stationary and non-stationary types of datasets. Performance evaluation on stationary datasets has exhibited that these online classifiers are equally capable as offline classifiers. Further, performance evaluation on non-stationary streaming datasets has exhibited that these classifiers are capable of handling large sized datasets under system memory constraint. The presented online one-class classifiers either outperformed existing online one-class classifiers (for most datasets) or yielded similar results (for some datasets). Overall, boundary framework based one-class classifiers (***ORK-OCELM(B)***) have performed better compared to reconstruction framework based one-class classifiers (***ORK-OCELM(R)***) as well as state-of-the-art online one-class classifiers.

As computational cost is the main concern in streaming data analysis, evaluation of the proposed classifiers verified that these are fast and computational efficient as compared to other classifiers. Hence, it can be stated that classifiers presented in this paper are a viable alternative to the existing one-class classifiers. In future work, fuzzy concept can be introduced with the presented methods for handling uncertainty present in the data.

ACKNOWLEDGMENT

This research was supported by Department of Electronics and Information Technology (DeITY, Govt. of India) under Visvesvaraya PHD scheme for electronics & IT. We would like to thank the editor and the anonymous reviewers that helped to greatly improve the quality of this manuscript.

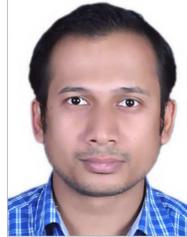

**Mr. Chandan Gautam** holds an M.Tech (IT) degree from University of Hyderabad and IDRBT, Hyderabad. He is currently a research scholar (PhD) at IIT Indore. His research interests include Noniterative Learning, Kernel Learning, Online Learning and One-Class classification. He is reviewer of various reputed journals like Neurocomputing, IEEE Transactions on Cybernetics, IEEE Systems Journal etc.

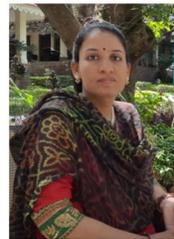

**Aruna Tiwari** received the B.E. degree (Computer Engineering) in 1994 and M.E. degree (Computer Engineering) in 2000 from SGSITS, Indore and Ph.D. degree from RGPV, Bhopal, India. She joined the Indian Institute of Technology, Indore, India, in 2012, where she is currently working as an Associate Professor with the Department of Computer Science and Engineering. Her research interests include soft computing techniques, neural network learning algorithms, genetic algorithms and evolutionary approaches. She has many publications in peer reviewed journals, international conferences, and book chapters. She is reviewer of many journals some of them are IEEE Transaction on KDE, Neurocomputing journal of Elsevier etc. Dr. Tiwari is life member of IEEE Computer Society of India and member of IEEE Evolutionary Computation Society.

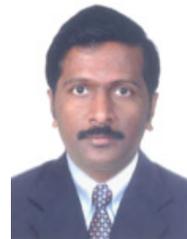

**Sundaram Suresh** received the B.E degree in electrical and electronics engineering from Bharathiyar University in 1999, and the M.E. and Ph.D. degrees in aerospace engineering from the IISc, Bangalore, India, in 2001 and 2005, respectively. He was a Post-Doctoral Researcher in the School of Electrical Engineering, NTU, Singapore from 2005 to 2007. From 2007 to 2008, he was with the INRIA-Sophia Antipolis, Nice, France as Research Fellow of the European Research Consortium for Informatics and Mathematics. He was with Korea University, Seoul, Korea, for a short period as a visiting faculty in Industrial Engineering. From January 2009 to December 2009, he was with the Department of Electrical Engineering, IIT Delhi, India, as an Assistant Professor. He is currently working as an Associate Professor with the School of Computer Engineering, NTU, Singapore. His research interest includes flight control, unmanned aerial vehicle design, machine learning, and optimization and computer vision.

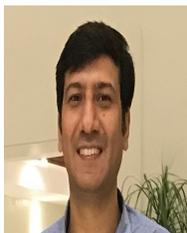

**Kapil Ahuja** (Bachelors: IIT (BHU), India; Masters: Virginia Tech, USA; PhD: Virginia Tech, USA; Postdoctoral Research Fellow: Max Planck Institute, Germany) has a varied background, including degrees in Computer Science, Mathematics, and Mechanical Engineering. He is currently an Associate Professor in Computer Science and Engineering at IIT Indore.

Dr. Ahuja works on applying mathematics and computation to solve science and engineering problems. Specifically, his research focuses on numerical linear algebra, numerical analysis, optimization, computational intelligence, big data, and social cloud.